\documentclass[conference]{IEEEtran}
\IEEEoverridecommandlockouts
\usepackage{cite}
\usepackage{amsmath,amssymb,amsfonts}
\usepackage{algorithmic}
\usepackage{graphicx}
\usepackage{textcomp}
\usepackage{xcolor}
\usepackage{balance}
\usepackage{fix2col}

\def\BibTeX{{\rm B\kern-.05em{\sc i\kern-.025em b}\kern-.08em
    T\kern-.1667em\lower.7ex\hbox{E}\kern-.125emX}}
\begin{document}

\title{Parallel Attention Mechanisms in Neural Machine Translation\\
}

\author{\IEEEauthorblockN{Julian Richard Medina}
\IEEEauthorblockA{\textit{Computer Science} \\
\textit{University of Colorado Colorado Springs}\\
Colorado Springs, CO, USA \\
jmedina5@uccs.edu}
\and
\IEEEauthorblockN{Jugal Kalita}
\IEEEauthorblockA{\textit{Computer Science} \\
\textit{University of Colorado Colorado Springs}\\
Colorado Springs, CO, USA \\
jkalita@uccs.edu}
}

\maketitle

\begin{abstract}
Recent papers in neural machine translation have proposed the strict use of attention mechanisms over previous standards such as recurrent and convolutional neural networks (RNNs and CNNs). We propose that by running traditionally stacked encoding branches from encoder-decoder attention-focused architectures in parallel, that even more sequential operations can be removed from the model and thereby decrease training time. In particular, we modify the recently published attention-based architecture called Transformer by Google, by replacing sequential attention modules with parallel ones, reducing the amount of training time and substantially improving BLEU scores at the same time.
Experiments over the English to German and English to French translation tasks show that our model establishes a new state of the art.
\end{abstract}


\section{Introduction}

Historically, statistical machine translation involved extensive work in the alignment of words and phrases developed by linguistic experts working with computer scientists \cite{Jurafsky2000}. Deep Learning surpasses these historically used methods and has primarily replaced these with the recent use of neural machine translation (NMT). The predominant design of the state of the art is the encoder-decoder model. The encoder takes sequential text, turning it into an internal representation. The decoder then takes this internal representation and generates a subsequent output. Since their emergence, attention mechanisms \cite{Bahdanau2014} have added to the effectiveness of the encoder decoder model and have been at the forefront of machine translation.

Attention mechanisms help the neural system focus on parts of the input, and possibly the output as it learns to translate. This concentration facilitates the capturing of dependencies between parts of the input and the output. After training the network, the attention mechanism enables the system to perform translations that can handle issues such as the movement of words and phrases, and fertility. 
However, even with these attention mechanisms, NMT models have their drawbacks, which include long training time and high computational requirements.

Recent papers \cite{Vaswani2017,Ahmed2017} in neural machine translation have proposed the strict use of attention mechanisms in networks such as the Transformer over previous approaches such as recurrent neural networks (RNNs) \cite{Elman1990} and convolutional neural networks (CNNs) \cite{LeCun1998}. In other words, these approaches dispense with recurrences and convolutions entirely. In practice, attention mechanisms have mostly been used with recurrent architectures because removing the recurrent nature of the architecture makes the training more efficient by the removal of necessary sequential steps.

This paper contributes by continuing to pursue the removal of sequential operations within encoder-decoder models. These operations are removed through the parallelization of previously stacked encoder layers. This new parallelized model can obtain a new state of the art in machine translation after being trained on one NVIDIA GTX 1070 for as little as three hours.

The paper includes the following: a discussion of related work in the field of machine translation including encoder-decoder models and attention mechanisms; an explanation of the proposed novel architecture with motivations; and a description of the used methodology, along with evaluation including used data sets, hardware, hyper-parameters, and metrics. This paper concludes with results and possible avenues for future research.
\section{Related Work}
There has been a plethora of work in the past several years on end-to-end neural translation. ByteNet \cite{Kalchbrenner2016} uses CNNs with dilated convolutions for both encoding and decoding. Zhou et al. \cite{Zhou2016} use stacked interleaved bi-directional LSTM layers (up to 16 layers) with skipped connections; ensembling gives the best results. Google's earlier and path-breaking end-to-end translation approach \cite{Wu2016} uses 16 LSTM layers with attention; once again, ensembling produces the best results. Facebook's end-to-end translation approach \cite{Gehring2017} depends entirely on CNNs with attention mechanism. 

Our work reported in this paper is based on another translation work by Google. Google's Vaswani et al. \cite{Vaswani2017} proposed the reduction in the sequential steps seen in CNNs and RNNs. The sole use of attention mechanisms and feed-forward networks within the common encoder-decoder sequential model replaces the necessity of deep convolutions for distant dependent relationships, and the memory and computation intensive operations required within recurrent networks. Original training and testing by Vaswani et al. were over both the WMT 2014 English-French (EN-FE) and English-German (EN-DE) data sets, while this paper uses only the WMT 2014 EN-DE set and the IWSLT 2014 EN-DE and EN-FR data sets. This model is discussed later in the paper.

Works in the field of NMT recommend a particular focus on the encoder. Analysis by Domhan \cite{Domhan2018} poses two questions: what type of attention is needed, and where. In this analysis, self-attention had a higher correspondence with accuracy when placed in the encoder section of the architecture than the decoder, even claiming that the decoder, when replaced with a CNN or RNN, retained the same accuracy with little to no loss in robustness. Imamura, Fujita, and Sumita's \cite{Imamura2018} study shows that the current paradigm of using high-volume sets of parallel corpora are sufficient for decoders but are unreliable for the encoder. These conclusions encourage further research in the manipulation of position and design of the encoder and attention mechanisms within them.

\section{Architecture}

The Transformer architectures proposed by Vaswani et al. \cite{Vaswani2017}, seen in Figure \ref{fig:BaseTransformer}, inspires this paper's work.  We have made modifications to this architecture, to make it more efficient. However, our modifications can be applied to any encoder-decoder based model and is architecture-agnostic. These alterations follow from the following two hypotheses.
\begin{enumerate}
    \item Reduction in the number of required sequential operations throughout the encoder section is likely to reduce training time without reducing performance.
    \item Replacing the subsequent encoder attention stack is expected to result in discarding of inter-dependencies, and possibly incorrect, assumptions of encoder attention mechanisms and layers, improving performance.
\end{enumerate}

For simplification, but without loss of generalization, this paper discusses the use and modification of Transformer based-models. The original Transformer model is composed of stacked self-attention layers. These self-attention mechanisms compare and relate multiple positions of one sequence in order to find a representation of itself.
In Figure \ref{fig:BaseTransformer}, we see such attention layers, one working on the input embedding, another on the output embedding, and the third on the both the input and the output embeddings. 
Each of these layers contains two main sub-layers including multi-head self attention, which feeds a simple feed-forward network, and a final layer of normalization. Around each of the main sub-layers, a skip or residual connection \cite{Kaiming2016} is also used. This same structure is used in the decoder with an attention mask to avoid attending to subsequent positions. 

The attention mechanism used by Vaswani et al. \cite{Vaswani2017} can be thought of as a function that maps a query and  set of key-value pairs to an output. The query, keys, values and output are all vectors. The output is obtained as a weighted sum of the values. The weight given to a value is learned by the system by considering how compatible  the query is to the corresponding key. The particular form of attention used is called {\em scaled dot-product attention}. This is due to the mechanism being homologous to a scaled version of the multiplicative attention proposed by Luong, Pham, and Manning \cite{Luong2015}. Several attention layers used in parallel constitute what is called {\em multi-head attention}.

A brief description the proposed modifications of this architecture is discussed below.

\subsection{Parallel Encoding Branches}

\begin{figure}[t!]

    \centering
    \includegraphics[height=4.5in]{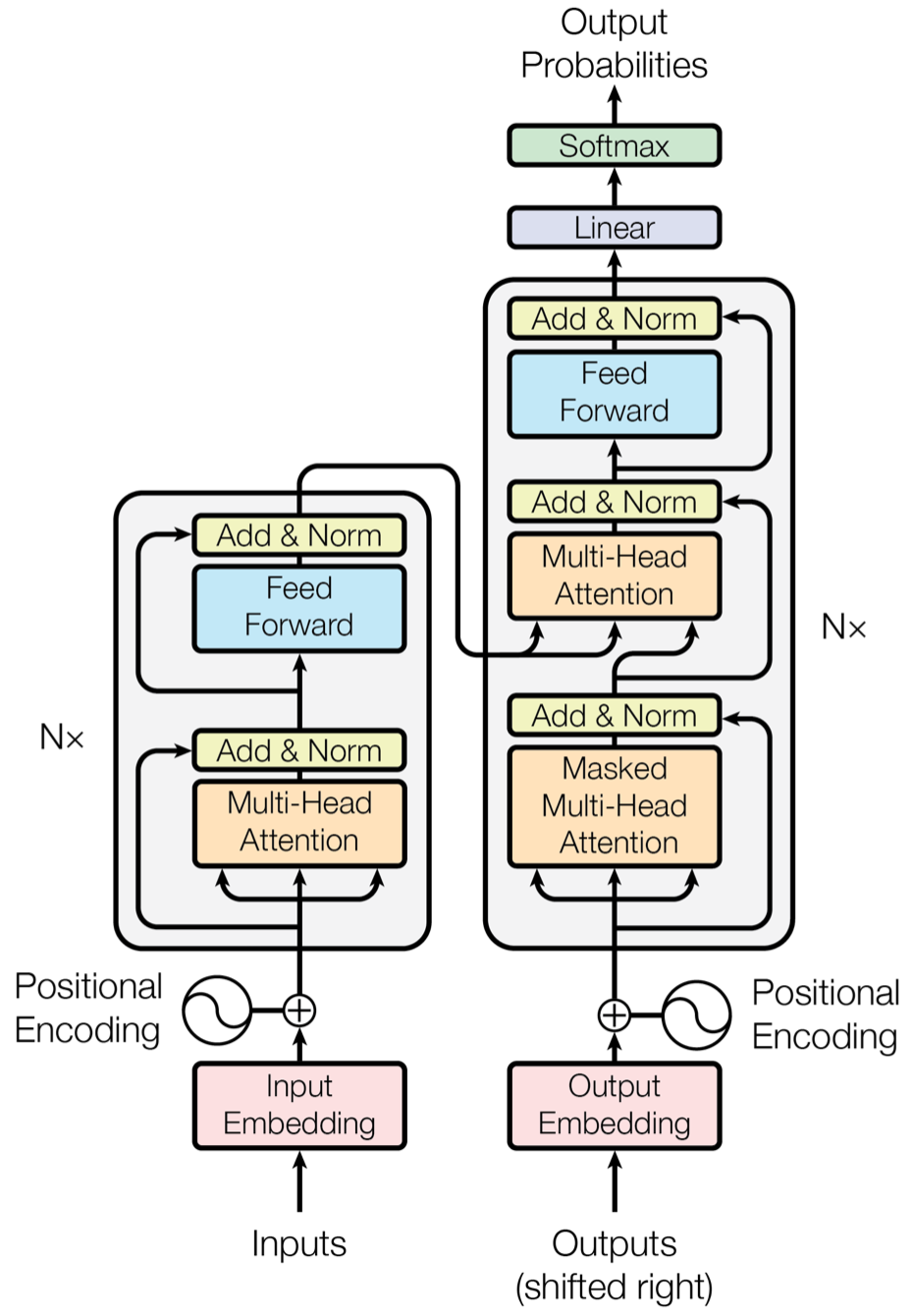}
    \caption{Transformer model as proposed by Vaswani et. al \cite{Vaswani2017}.}
    \label{fig:BaseTransformer}

\end{figure}

\begin{figure*}[h]

    \centering
    \includegraphics[height=3in]{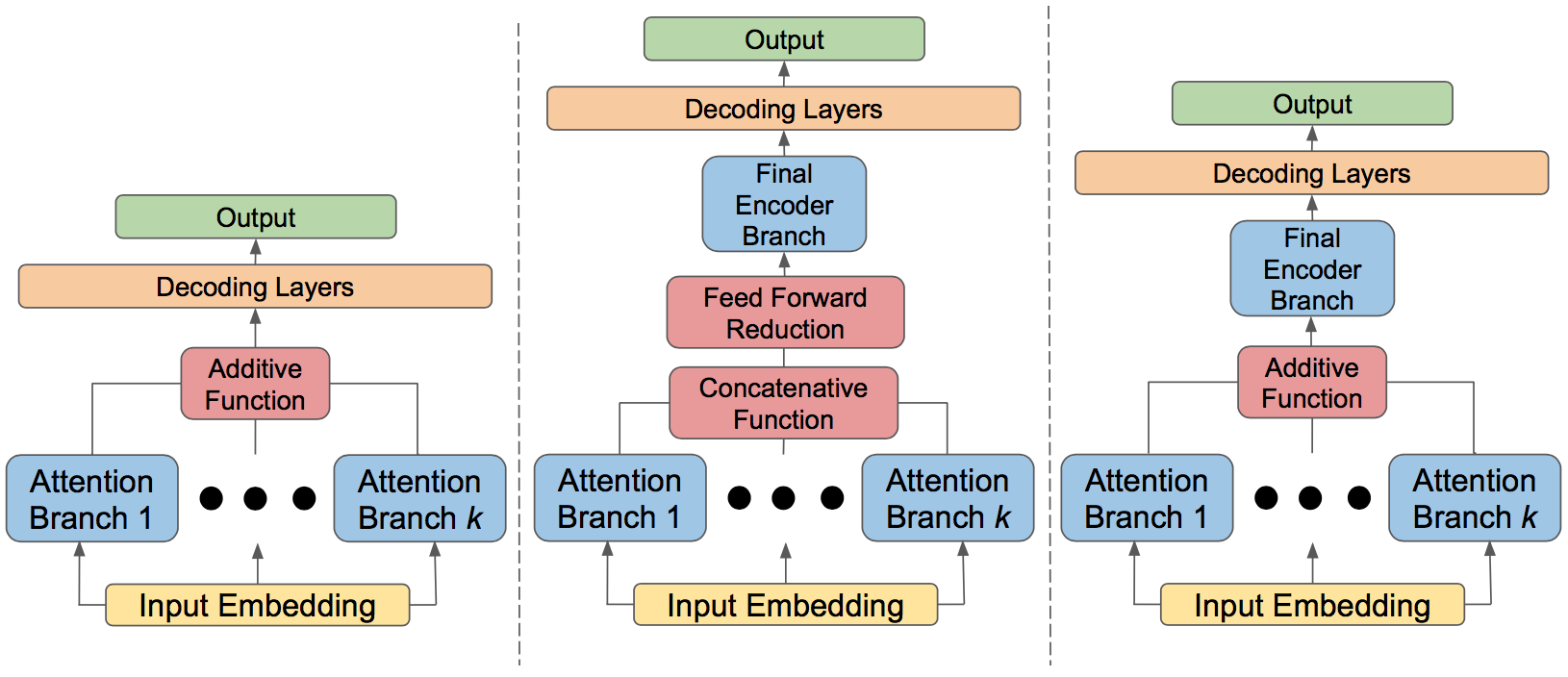}
    \caption{From left to right we present three models. 1) APA: Parallel encoded Transformer that uses homologous stacks of encoding trees with random initialization, and addition of their learned attention. 2) ACPA: Attended Parallel encoding where the branches concatenate learned results, a feed-forward network reduces dimensionality, and a final encoder branch encodes the results. 3) AAPA: Attended Parallel Encoding Branches where a final encoding attention branch attends to the added learned results.}
    \label{fig:AllModels}

\end{figure*}

\begin{table*}[t]
\begin{center}
\begin{tabular}{lccrr}
 & \multicolumn{2}{c}{No. Training Sentence Pairs} & \multicolumn{2}{c}{No. Testing Sentence Pairs}\\
Data Set & EN-DE & EN-FR & EN-DE & EN-FR\\
\hline
IWSLT \cite{Cettolo2012} & 197K & 220K & 628 & 622 \\
WMT & 4.5M & 36M & 3000 & 3000\\
\end{tabular}
\end{center}
\caption{Data sets used for training and testing for the translation tasks of English-German and English-French. The English-French statistics were included for the WMT data set although it is not directly used in the paper, as it will be included in future work.}
\label{table:DataSets}
\end{table*} 

A motivation for creating the Transformer model was the sluggish training and generation times of other common sequence-to-sequence models such as RNNs and CNNs \cite{Vaswani2017}. This was done by simplifying and limiting sequential operations and computational requirements while also increasing the model's ability to exploit current hardware architecture. This paper proposes that removal of the previously stacked branches of the encoder (there is a stack of $N$ encoder and other blocks on the left side of Figure \ref{fig:BaseTransformer}), parallelizing these separate encoder `trees', and incorporating their learned results for the decoder, will further eliminate sequential steps and accelerate learning within current sequence-to-sequence models. The architectures discussed are modeled in Figure \ref{fig:AllModels}.

Alterations to this parallel Transformer model were made and the following models were trained, tested, and are discussed in this paper:
\begin{itemize}
\setlength{\itemindent}{0.25in}
    \item Additive Parallel Attention (APA),
    \item Attended Concatenated Parallel Attention (ACPA), and
    \item Attended Additive Parallel Attention (AAPA).
\end{itemize}

\subsection{Model Variations}

\textit{Additive Parallel Attention (APA)}: We replace the entire stack of (multi-head attention, add and normalize, feed forward, add and normalize) repeated N times on the original Transformer architecture on the left column, on the input side. We instead have several such attention sub-networks in parallel. The output layers of these networks contain attention embeddings for the input. The values at the output layers among the stacks are added. This model is seen to the left in Fig. \ref{fig:AllModels}.

\textit{Attended Concatenated Parallel Attention (ACPA)}: This approach is similar to APA and AAPA, but the values at the output layers of the attention sub-networks are concatenated instead of being added. This model is seen in the middle of Fig. \ref{fig:AllModels}.

\textit{Attended Additive Parallel Attention (AAPA)}: This model is built similarly to the APA model. However, it removes one of the parallel stacks and uses it as a final sequential attention mechanism over the additive results. This model is seen to the right in Fig. \ref{fig:AllModels}.

When incorporating the results of the parallel encoding branches, two models of thought are pursued: additive and concatenation. The APA and AAPA models directly add the results of all encoding branches, whereas the ACPA models concatenate all encoding results and use a simple non-linear layer to learn a dimension-reduction among all attention branches. The attended parts of both the ACPA and AAPA models incorporate a final attention layer over all encoding branches before they are sent to the decoding layers.


\begin{table*}[t]
\begin{center}
\begin{tabular}{lccrr}
 & \multicolumn{2}{c}{BLEU} & \multicolumn{2}{c}{Single GPU Run-Time (s)}\\
Model & EN-DE & EN-FR & EN-DE & EN-FR\\
\hline
Transformer as proposed by Vaswani et al.\cite{Vaswani2017} & 47.57 \(\pm\) 4.97 & 56.15 \(\pm\) 0.42 & 8052.19 & 9480.70\\
Attended Additive Parallel Attention 5 Parallel Branches (AAPA) & \textbf{57.05} \(\pm\) 0.45 & \textbf{63.26} \(\pm\) 0.43 & 8158.26 & 9596.08\\
Attended Additive Parallel Attention 4 Branches & 56.22 \(\pm\) 0.63 & 62.68 \(\pm\) 0.25 & 7805.73 & 9114.84\\
Attended Additive Parallel Attention 3 Branches & 56.68 \(\pm\) 0.47 & 62.75 \(\pm\) 0.35 & 7412.81 & 8686.92\\
Attended Additive Parallel Attention 2 Branches & 55.94 \(\pm\) 0.01 & 61.24 \(\pm\) 0.53 & \textbf{6998.18} & \textbf{8228.14}\\
Attended Concatenated Parallel Attention (ACPA) & 48.67 \(\pm\) 4.47 & 62.31 \(\pm\) 0.21 & 8186.77 & 9710.70\\
\end{tabular}
\end{center}
\caption{Model comparison for test results over the \textbf{IWSLT 2014 test set}. The BLEU score is given as an average of the final epoch over multiple runs where also a standard deviation (SD) is also given. By reducing the number of parallel branches in the encoder, the model can maintain high accuracy and reduce run-time. All of these models were developed in in the OpenNMT toolkit \cite{Klein2017}.}
\label{table:ModelResults}
\end{table*} 

\begin{table*}[t]
\begin{center}
\begin{tabular}{lccrr}
Model & BLEU & Single GPU Run-Time (s)\\
\hline
Transformer Large \cite{Vaswani2017} & 60.95 & 168,806.61\\
Attended Additive Parallel Attention Large 7 Parallel Branches (AAPA) & 61.98 & 173,163.03\\
Transformer & 61.00 & 138,032.33\\
Attended Additive Parallel Attention 5 Parallel Branches & 62.69 & 141,041.74\\
Attended Additive Parallel Attention 4 Branches & \textbf{62.77} & 133,374.33\\
Attended Additive Parallel Attention 3 Branches & 62.07 & 123,929.10\\
Attended Additive Parallel Attention 2 Branches & 62.59 & \textbf{116,450.75}\\
Attended Concatenated Parallel Attention (ACPA) & 60.32 & 142,363.06\\
\end{tabular}
\end{center}
\caption{Model comparison for test results over the larger NMT English-German test set. All of these models were developed in in the OpenNMT toolkit \cite{Klein2017}.}
\label{table:ModelResultsBig}
\end{table*} 

\begin{table*}[t]
\begin{center}
\begin{tabular}{lccrr}
 & \multicolumn{2}{c}{BLEU}\\
Model & Cased & Uncased \\
\hline
Transformer Large \cite{Vaswani2017} & \textbf{24.20} \(\pm\) 0.081 & \textbf{23.72} \(\pm\) 0.005\\
Attended Additive Parallel Attention 3 Branches & 23.90 \(\pm\) 0.04 & 23.406 \(\pm\) 0.04\\
Attended Additive Parallel Attention 2 Branches & 23.794 \(\pm\) 0.28 & 23.494 \(\pm\) 0.02\\
\end{tabular}
\end{center}
\caption{Model comparison over the \textbf{WMT 2016} English-German translation task with our models implemented in the Tensor2Tensor \cite{Vaswani2018} library by Google. Each model was trained to 250K training steps. Although our models are comparable to the Transformer model for the WMT EN-DE task, they surpass Transformer for the IWSLT 2014 test set.}
\label{table:ModelResultsTensor2Tensor}
\end{table*} 
\section{Experiments and Evaluation}

All proposed architectures including the base Transformer model \cite{Vaswani2017} are trained over the International Workshop on Spoken Language Translation (IWSLT) 2016 corpus and tested similarly over the IWSLT 2014 test corpus \cite{Cettolo2012}. The training corpus includes over 200,000 parallel sentence pairs, and 4 million tokens for each language. The testing set contains 1,250 sentences, and 20-30 thousand tokens for French and German. This paper also performed experiments over the larger WMT data set including 4.5 and 36 million training sentence pairs for the EN-DE and EN-FR tasks respectively. The testing set for these experiments was the standard Newstest 2014 test set including around 3000 sentence pairs for each language task. These statistics are noted in Table \ref{table:DataSets}. The sentence pairs range in length from one to sixty tokens to get a full measure of the tested models and robustness to both short and long input.

Across all models, a greedy-decoding function for both training and testing time, the Kullback-Leibler divergence loss function, the Adam optimizer \cite{Kingma2014}, and the number of training epochs (10) were kept constant. The training and testing were done using the NMT task of English to German (EN-DE)  and IWSLT English to French and English to German translation and each network was trained using one graphics processing unit (GPU). The utilized machine GPU configuration was one NVIDIA GTX 1070.

\begin{figure}[h!]

    \centering
    \includegraphics[height=1.5in]{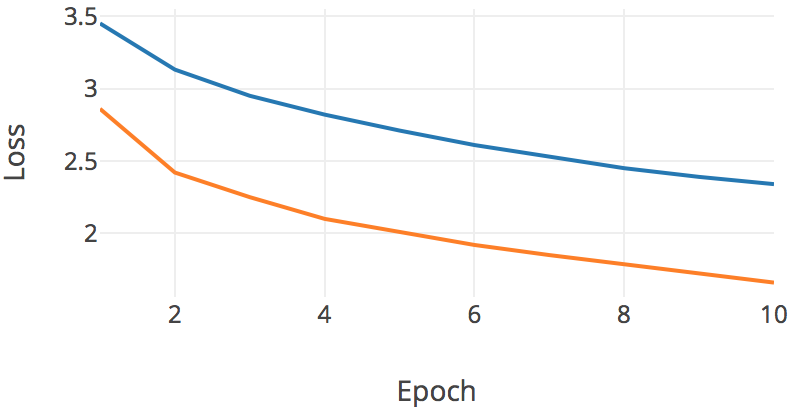}
    \caption{This plot shows validation loss for both the Transformer model (blue) and our modified model (orange) over the IWSLT EN-DE task. The parallel encoder shows a consistently lower starting and end-training loss.}
    \label{fig:LossEpoch}

\end{figure}

\begin{figure}[h!]

    \centering
    \includegraphics[height=1.5in]{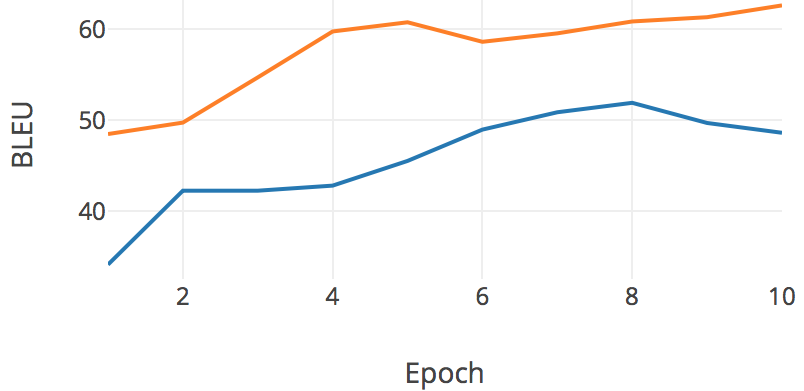}
    \caption{This plot shows validation BLEU metric score for both the Transformer model (blue) and our modified model (orange) over the IWSLT EN-DE task. The parallel encoder shows a consistently higher BLEU score and shows linear increase while the Transformer shows some plateauing in later epochs.}
    \label{fig:BLEUEpoch}

\end{figure}

For the assessment of each model and translation task this paper uses the bilingual evaluation understudy (BLEU) metric \cite{Papineni2002}. This is a modified precision calculation using n-grams such as unigram, grouped unigrams, and bigrams. The BLEU metric claims to have a high correlation to translation quality judgments made by humans. BLEU computes scores for individual sentences by comparing them with good quality reference translations. The individual scores are averaged over the the entire corpus, without taking intelligibility or grammatical correctness into account.

\begin{figure*}[!h]

    \centering
    \includegraphics[height=5in]{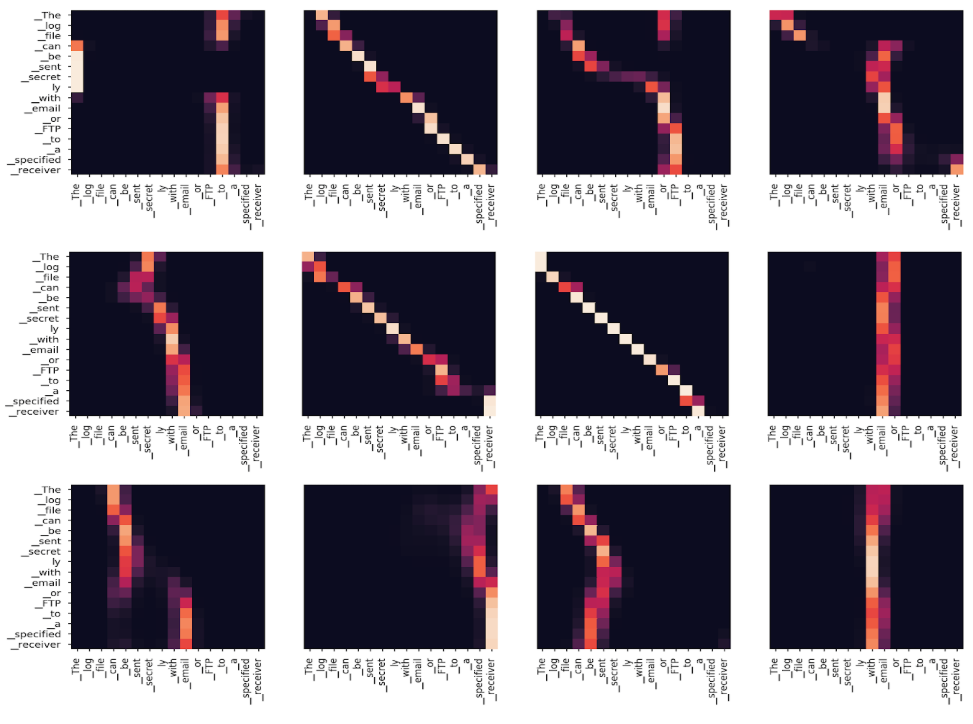}
    \caption{Visualization of multi-head attention weights in encoder branches 0, 2, and 4. Although each receives the same input embedding, through random initialization, each learns different focuses.}
    \label{fig:AttentionVisualization}

\end{figure*}

\begin{figure*}[!h]

    \centering
    \includegraphics[height=1.9in]{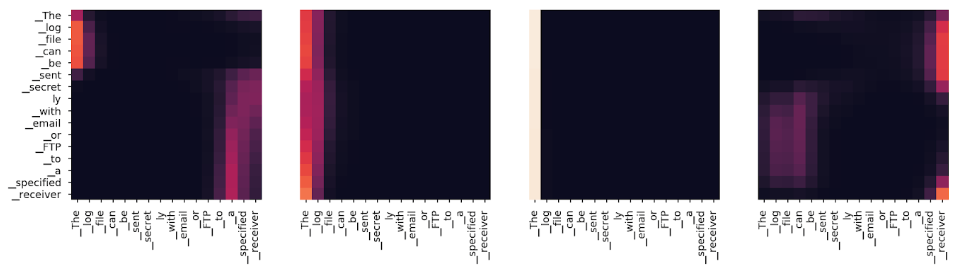}
    \caption{Visualization of the weights for the final encoder that attends over all other encoding branches. This encoder's weights are relatively light, abstract, and have less obvious patterns when compared to the individual encoding branches.}
    \label{fig:FinalAttentionVisualization}

\end{figure*}
\section{Results}

\subsection{Attention Visualization}

One concern during early hypothesis testing was that if each attention branch looks at the same input, that each one would learn to focus on the same properties of the original embedding. However, through visualization of each attention layer, it is obvious that regardless of the same input, the encoder branches through random initialization learn different focuses as seen in Figure \ref{fig:AttentionVisualization}. The final branch for the attended models however would learn very light to no attention weights as seen in Figure \ref{fig:FinalAttentionVisualization}. This is one area of research this group wishes to pursue in the future.

\subsection{Machine Translation}

Table \ref{table:ModelResults} shows that the AAPA model consistently performed on average nearly ten points higher in the BLEU metric on the English to German translation task on the IWSLT 2014 test set. It also performed very well on the English to French translation task.

On the much larger WMT English-German test set, all our models achieve better results then Vaswani et al. \cite{Vaswani2017}. Our model with five parallel encoding branches has a BLEU score of 62.69 compared to 60.95 and 61.00 for the two Transformers shown in Table \ref{table:ModelResultsBig}. Our approach also takes considerably less time than the large Transformer model with a stack of eight encoder attention heads, although it is a little slower than the smaller Transformer model reported by Vaswani et al. \cite{Vaswani2017}. In terms of the BLEU metric, we establish state-of-the-art performance for both EN-DE and EN-FR translation considering the IWSLT 2014, and comparable results for the WMT data sets.
Since our results came up very good, surpassing state of the art for the IWSLT 2014 dataset, we ran our experiments multiple times to ensure the results are correct. 

During the Transformer and attended parallel model's training lifetime, it can be seen that loss was consistently lower for our modified parallel model with five parallel stacks as seen in Figure \ref{fig:LossEpoch}. In this task, loss doesn't always correspond to a higher metric, in this case our model also shows a continuous higher score in the BLEU metric over the validation set while the Transformer shows signs of plateauing early on Figure \ref{fig:BLEUEpoch}.

However, our parallelized model did have a slightly higher training time over a single GPU. One final experiment conducted to improve this drawback, also seen in the same table, is the reduction of number of parallel branches in the encoder. By reducing the number incrementally, our BLEU score stays equivalent to higher perplexity layers, but linearly reduces the run-time.

\section{Conclusions}

In step with the goals of the original Transformer, this work continued to pursue the removal of sequential operations within attention-based translation models. Although dependent on choice of tool-kit implementation as shown in Table \ref{table:ModelResultsTensor2Tensor}, this new parallelized Transformer model reaches a new state-of-the-art in machine translation and provides multiple new directions for future research. It also shows through random initialization that attention mechanisms can learn different focuses and that by eliminating possibly negative inter-dependencies among them, superior results can be obtained.

\section{Acknowledgement}

This material is based upon work supported by the National Science Foundation under Grant No. 1659788.  Any opinions, findings, and conclusions or recommendations expressed in this material are those of the author(s) and do not necessarily reflect the views of the National Science Foundation.

\balance

\end{document}